# A Neuromorphic Paradigm for Online Unsupervised Clustering

J. E. Smith
University of Wisconsin-Madison (Emeritus)
Carnegie Mellon University (Adjunct)

## ABSTRACT

A computational paradigm based on neuroscientific concepts is proposed and shown to be capable of *online unsupervised clustering*. Because it is an online method, it is readily amenable to streaming realtime applications and is capable of dynamically adjusting to macro-level input changes. All operations, both training and inference, are localized and efficient.

The paradigm is implemented as a cognitive *column* that incorporates five key elements: 1) temporal coding, 2) an excitatory neuron model for inference, 3) winner-take-all inhibition, 4) a column architecture that combines excitation and inhibition, 5) localized training via spike timing dependent plasticity (STDP). These elements are described and discussed, and a prototype column is given. The prototype column is simulated with a semi-synthetic benchmark and is shown to have performance characteristics on par with classic *k*-means. Simulations reveal the inner operation and capabilities of the column with emphasis on excitatory neuron response functions and STDP implementations.

## 1 INTRODUCTION

The many-decades effort to understand, and then replicate, the brain's computational paradigm(s) is far from complete. Although neuroscientific experiment and theory have revealed much about the elements of neural computation, refining and combining them into a cohesive, widely accepted paradigm remains the subject of intensive ongoing research.

One line of research targets deep, hierarchical spiking neural networks (SNNs) [53], similar in structure to convolutional neural network (CNN) classifiers. Much of the SNN research targets improved energy efficiency when implementing supervised classification. In contrast, the research reported here does not address problems for which state-of-the-art machine learning already excels. Rather, the goal is to tackle a problem for which conventional machine learning methods are less adept, but for which neuromorphic methods appear to be well-suited: *online unsupervised clustering*.

*Clustering* partitions a set of input patterns into groups where the members of a cluster are more similar to each other than to members of other clusters. An *online* implementation consumes and processes inputs item-by-item; there is no buffering of inputs for deferred processing. This feature supports realtime processing of streaming inputs in a natural way. Because clustering is achieved online without metadata, if input data patterns change at a macro-level, then the clustering function adapts by dynamically reformulating the clusters.

This paper demonstrates that online unsupervised clustering is achievable via a simple, neuromorphic paradigm that can be described as a centroid-based clustering method. Spike Timing Dependent Plasticity (STDP) is at the core of the online clustering mechanism. STDP operates independently at each synapse, using only locally available information. Consequently, learning is highly parallel, fast, efficient, and displays the emergent behavior one would expect of a brain-like learning mechanism.

### 1.1 Application Domain: Edge Processing

Online unsupervised clustering is a kernel function for many important edge-processing tasks. Examples include: 1) removing noise from sensor data, 2) compressing sensor information prior to sending it to a host process, thereby reducing transmission energy, 3) pre-processing data, thereby reducing the amount of AI processing performed at the host, 4) detecting anomalous behavior and triggering a supervisor process that intervention is required.

A large fraction of the brain's neocortex is devoted to sensory processing, and much of what is known about neocortical function comes from *in vivo* study of sensory processing. Consequently, it makes sense to consider neuromorphic methods that are suitable for edge processing tasks. And, just as the neocortex processes sensory information in a hierarchical manner, so might the development of neuromorphic methods proceed up a hierarchy, beginning with close-to-the edge functions, such as the one studied in this paper.

The long term objective of this research is highly efficient, flexible, adaptable edge processing hardware. However, the focus here is on the underlying paradigm. So, even though this paper is focused on function, an eventual direct hardware implementation drives modeling decisions.

### 1.2 The Column

A *column* is a functional block that performs online unsupervised clustering. The term "column" is chosen because the scale and computational capabilities are roughly the same as a biological column in the neocortex [43]. However, it is not suggested that the internal organization and interconnections are the same. At this point, the similarity goes no further than the level of abstraction, the scale, and, it is hypothesized, the basic function being implemented.

Studying a standalone column is of primary interest here, but a topic of ongoing research is the development of *Temporal Neural Networks* (TNNs) consisting of multiple, hierarchical columns. Often in the literature, TNNs that employ temporal encoding are referred to as the more generic "SNN". However, not all SNNs are TNNs. Many proposed SNNs use spike rates to encode values, rather than relationships among individual spike times as in TNNs.

### 1.3 Contributions

Online unsupervised clustering can be a key enabler of energy efficient edge processing applications, *provided* it can be implemented in a way that is simple, accurate, and has low online training costs.

At a high level, the paper does three things: 1) Establishes a paradigm that can be described and easily understood in terms of centroid-based operations, 2) Proposes a prototype implementation that combines a number of concepts from neuroscience, 3) Includes simulation results that demonstrate the basic implementation works as the centroid-based rationale says it should. Along the way significant contributions to TNN theory are made:

1) STDP is a central topic. Classically, STDP is considered for cases where synaptic weight updates are based on the occurrence of both an input and an output spike. However, it is demonstrated here that cases where there is only one spike are at least as important. Also, an STDP search mode is shown to reduce learning times and permit arbitrary initial synaptic weights. Consequently, there is good support for online adaptability.

2) It is shown that neuron response functions with a sloping leading edge (a ramp) provide more functionality than a step leading edge. A trend today is toward step response functions, at least partly because they are readily amenable to event-driven implementations and simulation. However, this sacrifices temporal capabilities offered by a ramp, so the approach here runs counter to the trend.

3) It is based on a *small integer* model rather than a commonly-used *real* model. From the bottom-up, this approach supports a simple direct hardware implementation.

## 2 Background and Related Research

The proposed column architecture is based on five main elements drawn from the neuroscience literature: 1) Temporal coding 2) Excitatory neuron model (inference) 3) Inhibition model 4) Column architecture 5) STDP training. These are each discussed in historical sequence.

### 2.1 Temporal Coding

In 1982, Abeles proposed a high level model for explaining neocortical computation [1][2]. *Synchronous synfire chains* are groups of excitatory neurons, organized into layers, that process and pass information as a wave, or volley, of precisely timed spikes.

In 1989, Thorpe and Imbert [55] made a persuasive experimentally-supported argument for inter-neuron communication via precisely timed spikes. In contrast, communication via rate encoding is implausibly slow. In 1990 Thorpe [56] proposed precise spike timing relationships as a basis for neocortical communication. Strongly stimulated neurons produce relatively early spike times, and less strongly stimulated neurons yield later spike times.

In 1995, Hopfield [28] proposed a specific temporal coding method (Figure 1) where synaptic weights control neuron input delays. Training the weights essentially tunes the delays to temporal spike patterns. Hopfield further proposed that model neurons exhibit radial basis function (RBF) behavior, where output spike timing indicates how well a given input pattern matches the pattern of trained weights.

Regarding precision, Gray et al. [23] suggest that spike volleys are synchronized by gamma oscillations, leading to a temporal coding interval of 5-10 msec. within which all spikes in a coordinated volley occur. Experiments show that excitatory neuron spiking behavior is repeatable to within 1 msec [11] [37]. Combining a 5-10 msec coding window with 1 msec coding precision yields about 3 bits of resolution (4 at most). Consequently, the computing model developed in this paper is based on low resolution integers.

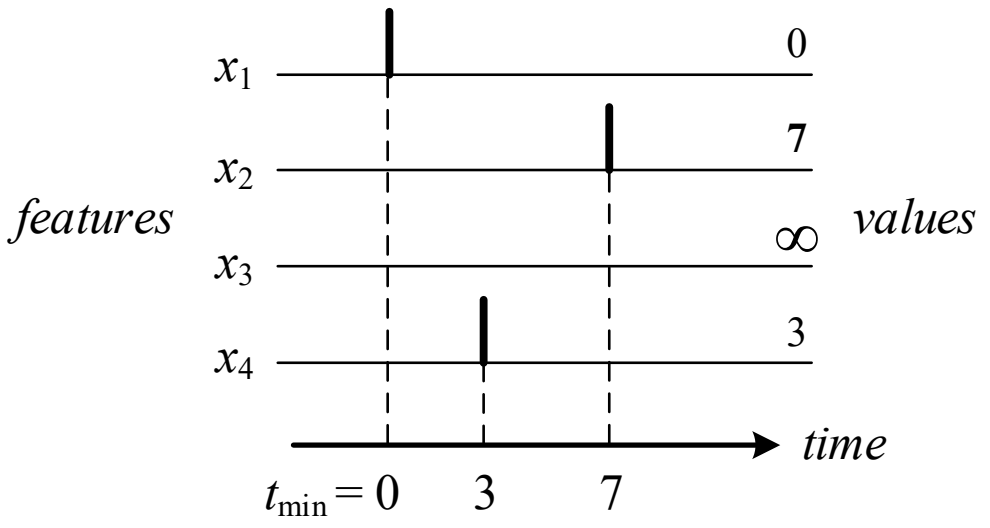

**Figure 1. A spike volley. The presence of a spike indicates the presence of a feature and the time of a spike indicates the feature's relative strength, with 0 being the strongest.**

### 2.2 Excitatory Neuron Model (Inference)

Seminal work in neuron modeling is due to Hodgkin and Huxley [27]. Their objective was a high degree of biological accuracy. Toward the opposite end of the complexity scale lie models emphasizing computation, many of which are variants of a "leaky-integrate-and-fire" (LIF) model [20][52][57].

An elegantly simple, and widely used, model for neuron behavior is the Spike Response Model (SRM) proposed in 1994 [21]. A simple form of SRM is illustrated in Figure 2a. In the SRM0, input $x_i$ connects to the neuron body via synapse $i$ having weight $w_i$. The neuron operates as follows.

1) If there is a spike on $x_i$, then the value of the synaptic weight $w_i$ selects a pre-defined *response function*.

2) In the neuron body, the synaptic response functions are summed, yielding a net *body potential*.

3) When, and if, the body potential reaches a *threshold* value θ, a spike on output $z$ is produced at that time.

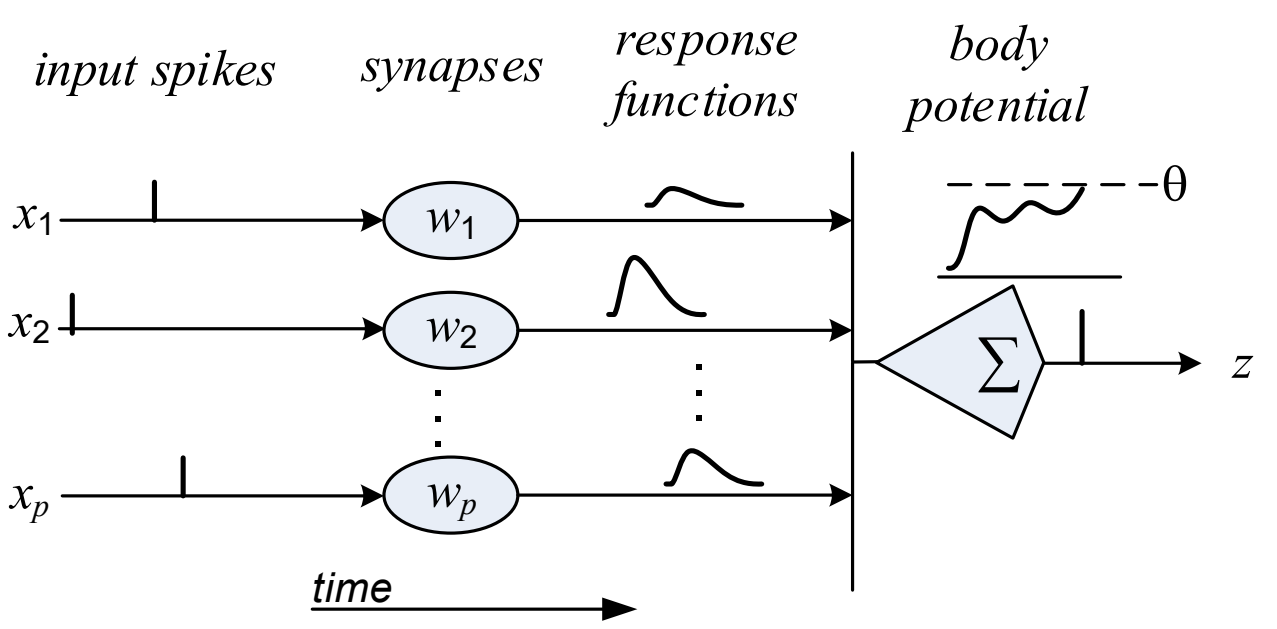

**a) SRM0 neuron operation**

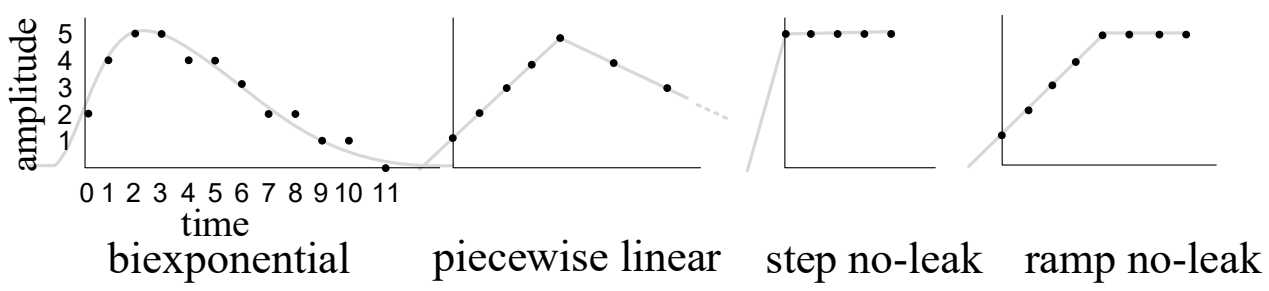

**b) Four example response functions**

**Figure 2. SRM0 excitatory neuron model.**

Typically, different synaptic weights map to response functions that differ only in amplitude, although this is not a requirement. Example response functions are in Figure 2b. The Tempotron model [24] uses a *biexponential* response function. Maass [36] uses *piecewise linear* response functions. Some more recent research uses *step no-leak* [18][19][31]. The response function used in this paper is *ramp no-leak.*

In the mid-1990s, Maass [36] used a spike response model for studying the computational capabilities of spiking neurons, in which precise spike timing relationships are fundamental to the model. Maass defines temporal coding and TNNs formally. (he calls them "SNNs", but as noted earlier, the scope of "SNNs" has since expanded to include rate-based coding methods).

### 2.3 Winner-Take-All Inhibition

Winner-Take-All (WTA) methods go far back in the annals of machine learning, and in one form or another WTA is ubiquitous amongst TNNs. An important experimentally-supported principle is that, unlike excitation, inhibition is a bulk process where inhibitory neurons act collectively to lay down a "blanket of inhibition" [29]. The WTA function is invoked by the first (strongest) spike in a volley and prevents other excitatory neurons from spiking.

WTA inhibition over a bundle of lines was proposed by Thorpe in 1990 [56] and was used by Natschläger and Ruf in their seminal 1998 paper [44]. There have been several variations of WTA, for example $k$-WTA where the $k$ first spikes are allowed to pass through uninhibited [38].

### 2.4 Column Architecture

An early column architecture of the type considered in this paper was proposed and studied by Natschläger and Ruf [44]. See Figure 3. Their objective is "RBF behavior" achieved by combining excitatory neurons with WTA inhibition. A parallel group of $q$ excitatory neurons are trained without supervision to identify $q$ RBF-like centroids. Then, inference proceeds as follows: 1) an input is applied, 2) each RBF neuron implicitly computes the distance with respect to its centroid, and 3) WTA inhibition selects the first neuron to spike; i.e., the one that computes the least distance.

Early efforts by Simon Thorpe and his group combined a form of STDP with "rank-order coding" [58][60] that used feedforward inhibition as a mechanism for encoding temporal relationships. A later, more conventional STDP method, akin to the one used in this paper, is described in [25], and a full column architecture is in [40]. It uses a no-leak response function because, as the authors argue, the leak observed in biology is merely a re-set mechanism, which can be simulated in other, simpler ways. A no-leak response function is used in this paper for the same reason.

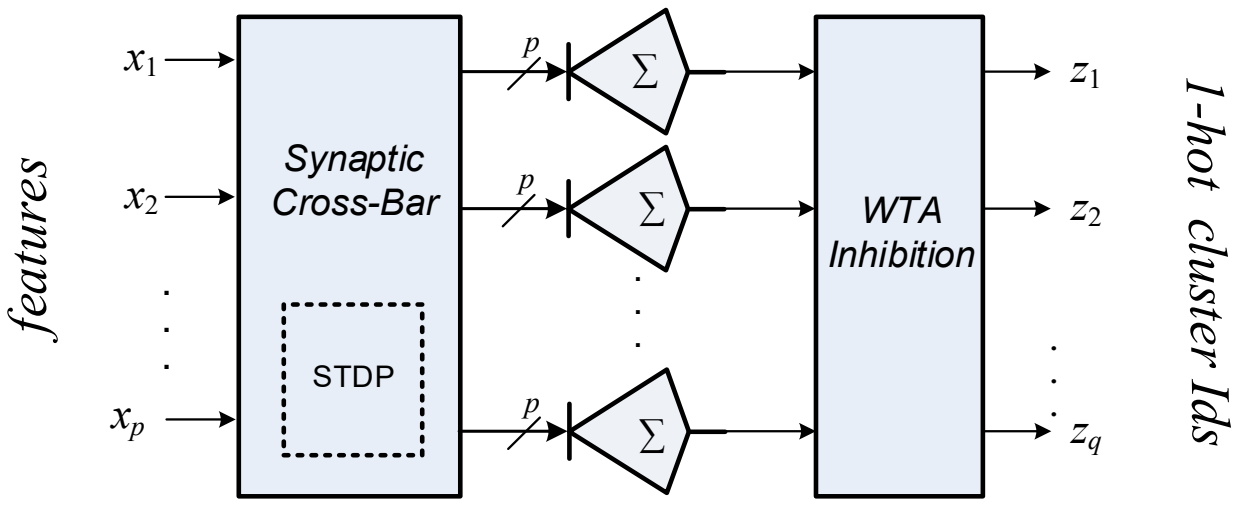

**Figure 3. A column consists of a synaptic crossbar feeding a column of parallel excitatory neurons. WTA inhibition selects the first (strongest) output. STDP adjusts weights according to input and output spike times.**

### 2.5 STDP

Hebb [26] observed that repeated temporal coincidence between a synapse's input spike and its neuron's output spike tends to increase the synapse's weight. Although insightful, this observation is also very general. In 1983, Levy and Steward [35] established the classic STDP update rules: if an input spike precedes its neuron's output spike, then the synaptic weight is increased; if the input spike follows its neuron's output spike, then the synaptic weight is decreased.

Gerstner et al. [22] proposed a theoretical foundation for STDP as an integral part of a computing paradigm. The key feature is weight update rules that result in a "competitive self-organized process" that trains a neuron's synapses to recognize certain input spiking patterns. This proposition was supported by independent experimental work by Markram et al. [39] . Bi and Poo [6] provided additional experimental support and fleshed out a more complete functional model for STDP.

### 2.6 Spiking Neural Networks

As considered in this paper, a TNN embodies *all five* of the key elements just listed. Most of the research on SNNs, however, employs only a proper subset, and therefore differs significantly from the approach taken here.

Some SNNs use rate-coding, not temporal coding [12][29][32][34][45][50][51]. Values are expressed as spike rates, typically Poisson-encoded. Most of these networks train with back propagation, often using schemes transported or transformed from CNNs. Other SNNs combine rate coding with STDP [5][10][16][17] [46][54] or with an auto-encoder method [48] that achieves unsupervised learning. Other SNNs use temporal coding and train via back propagation [9][14][42][49][61]. As with rate-coded SNNs, back propagation is typically transported or transformed from conventional CNNs.

The above covers a large fraction of SNNs, leaving TNNs that employ all five features (or at least come close). RBF neural networks[8][44][41] have all the features, although the STDP method is unorthodox, as is the synaptic network. The Tempotron [24] uses supervised STDP to construct centroids based on labels.

A notable body of TNN research comes from Thorpe et al. [7][25][31][40][58]. This model is also used in [18][59]. The research in this paper is mostly closely aligned with this body of research. The approach in [31] is typical and features of that work are singled out later for comparisons.

### 2.7 Platforms

Some SNN research is directed at "platforms" rather than "paradigms". Platforms support spiking neuron models, but are not computing models themselves. They have lots of flexibility designed-in so they may support research on a wide variety of models.

Platforms range from simulators to special-purpose hardware. Most closely related to research in this paper is special-purpose hardware constructed with conventional CMOS. The IBM TrueNorth system is a prime example [4][13]. The Intel Loihi [15] is a more recent effort.

Exploring direct CMOS implementations is a long range objective of this research effort. However, the approach taken here is to first develop and optimize a computing model via software simulation. Then, a direct application-driven CMOS implementation will be considered. The eventual CMOS implementation will lack much of the flexibility of the research platforms, but will not be burdened with the hardware and efficiency overheads that come with flexibility.

## 3 Column Architecture

An overall block diagram is in Figure 3. The column architecture operates on an input *volley* $x^i$ consisting of $p$ *spikes*: $x^i = [x^i_1, \ldots\ x^i_p]$, $x^i_j \in \mathrm{N}_0^\infty$. The superscripts are omitted when there is no ambiguity. The set $\mathrm{N}_0^\infty$ models time in discrete units and consists of the non-negative integers plus the special symbol "∞" that models the case where there is no spike. In Figure 1, $x = [0, 3, \infty, 1]$.

An input volley, encodes a *vector*, *pattern* or *image* (these will be used interchangeably). A spike indicates the presence of an associated *feature*, and the relative strength of the feature is encoded as the spike time relative to the first in a volley. I.e., an earlier spike indicates a stronger feature. A feature can be as simple as a single pixel.

Via an unsupervised training process, a column partitions a multi-set of patterns $\mathrm{P} = [x^1, x^2, \ldots, x^{|\mathrm{P}|}]$ into *clusters* of similar patterns:

$$\mathrm{C}_i \subseteq \mathrm{P}\ ;\ \forall\, i, j: i \neq j\ \mathrm{C}_i \cap \mathrm{C}_j = \emptyset\ .$$

As commonly defined, the *centroid* $\mathrm{c}^k$ of cluster $\mathrm{C}_k$ is an element-wise average of all the members of $\mathrm{C}_k$.

Online unsupervised clustering is performed by passing input volley $x$ through a $p \times q$ synaptic crossbar to $q$ SRM0 neurons. The crosspoints hold a weight matrix $W$; $0 \leq w_{ij} \leq w_{\max}$. $W$ is established via a column-level unsupervised training process that adjusts the weights so that each neuron eventually identifies a cluster i.e., there is a high correlation between a neuron's synaptic weights and a cluster's centroid. (Note that the learned "centroids" are only approximations and are not strictly-defined mathematical centroids.) For a given input $x$, the excitatory neurons evaluate centroid distances (approximately), and the neuron associated with the nearest centroid fires first. Subsequently, WTA inhibition selects the nearest centroid, say $\mathrm{c}^k$, and a spike on output $z_k$ serves as a cluster identifier.

Furthermore, the spike time on output $z_k$ indicates the relative *distance* from the nearest centroid. Hence, the inference process exhibits "RBF behavior" [44] by determining 1) cluster membership, indicated by the presence of a spike, and 2) the distance from the cluster's centroid, indicated by the spike's relative time.

### 3.1 Excitatory Columns (Inference)

In this section, it is assumed that training has established a stable set of weights. To simplify discussion, assume a direct relationship between times and values (see Figure 1). If a temporal event occurs at time $t$, then its value is $t$.

Excitatory neurons employ *ramp no-leak response functions* $\rho(w,t)$ that map an integer weight $0 \leq w \leq w_{\max}$ and an integer time $t$ onto the non-negative integers.

$$\begin{aligned} \rho\,(w,t) &= 0 && \text{if } t < 0 \\ &= t + 1 && \text{if } 0 \leq t < w \\ &= w && \text{if } w \leq t \end{aligned}$$

Neuron $j$ generates output $y_j$ as a two-step process. First, spike-time-shifted response functions are summed at the body of neuron $j$ to yield a *body potential*, $v_j$:

$$v_j(t) = \Sigma\, \rho(w_{ij}, t - x_i) \quad \text{for } i = 1..p$$

Second, depending on the body potential, the output is produced via the *spiking function* $\sigma$:

$$y_j = \sigma(v_j(t), \theta) = \text{the smallest } t \text{ for which } v_j(t) \geq \theta;$$

$$y_j = \infty \ \text{ if there is no such } t.$$

Collectively, $q$ neurons perform function E that produces a $q$ element output vector $y$. I.e., $y = \mathrm{E}(x,W)$. It is assumed that during operation the threshold $\theta$ is fixed and therefore is not included as a function input.

### 3.2 Inhibition

In this paper, one of the simpler forms of inhibition is implemented: 1-WTA. Only the input spike with the earliest spike time is passed through as an output spike. If there is a tie for the earliest spike, the tie is broken systematically by using the lowest index.

As shown in Figure 3, WTA inhibition acts on volleys of size $q$, so the inhibitory function I has input lines $y_{1..q}$ and output lines $z_{1..q}$. Define $y_{\min} = min(y_i)$. Then $z = \mathrm{I}(y)$, where

$z_i = y_i$ if $y_i = y_{\min}$ and for any $k < i$ , $y_k \neq y_{\min}$

$z_i = \infty$ otherwise.

### 3.3 Inference

The inference process determines cluster membership by evaluating the set of vector functions $z = \mathrm{I}(\mathrm{E}(x,W))$ which define a set of $q$ clusters. Pattern $x \in \mathrm{C}_i$ iff there is a spike on output $z_i$, i.e., $z_i \in \mathrm{N}_0$. Note that it is not required that every pattern $x$ belongs to a cluster; i.e., for some patterns, $z_i = \infty$ for all $i$.

The definition of inference is quite concise. Following is a lengthier discussion of the underlying concepts and rationale, using a simplified example.

Of primary interest in this paper is temporal processing, consequently, although it may seem odd, primary inputs are temporally flattened by *binarizing* them. That is, all input lines with a spike ($x_i \neq \infty$ ) are converted to $x_i = 0$. For example: *binarize*([0, 3, ∞, 1]) = [0, 0, ∞, 0]. After binarizing, a spike indicates the presence or absence of a feature, not its relative strength. Although column *inputs* are binarized, column *outputs* are not. From an experimental perspective, the idea is that by removing temporal information from the inputs, all temporal information in the outputs must be due to temporal processing. This gives clearer insight regarding the features and capabilities of temporal processing; in particular, it exposes RBF behavior.

The STDP strategy strives for weights that stabilize at a bimodal distribution; i.e., the predominant weights are 0 and $w_{\max} = 8$. The weights associated with a given neuron's inputs are strongly correlated with its cluster's centroid in the following way: a weight of $w_{\max}$ indicates a feature more likely to belong to members of the cluster and a weight of 0 indicates a less likely feature.

Because binarized input spikes all occur at $t = 0$, a neuron's body potential is the sum of ramp no-leak response functions with no time shifts: $v(t) = \Sigma\, \rho(w_i, t)$. An input $x$ containing $k$ spikes feeding synapses with weight $w_{\max}$ yields $v(t) = k\, \rho(w_{max}, t)$. Examples are plotted in Figure 4. Body potential for 1 through 4 spikes, where all syn-apses have weight $w_{\max} = 8$. for $k = 1..4$. For $\theta = 8$, spike times are shown with dashed arrows. The output spike times carry information regarding the relative strength of the associated feature match -- or the relative distance from the centroid. Given this information, WTA inhibition selects the output with the lowest value, indicating the most feature matches.

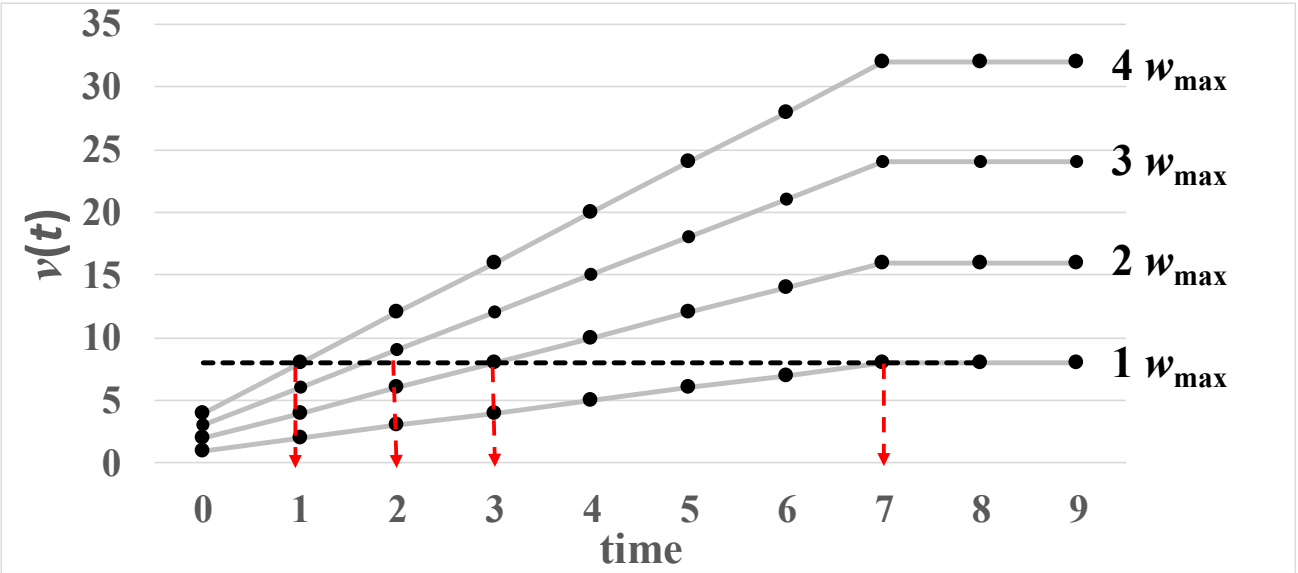

**Figure 4. Body potential for 1 through 4 spikes, where all synapses have weight $w_{\max} = 8$. For threshold $\theta = 8$, the output spike times are given on the x axis.**

### 3.4 STDP

The objective of Spike Timing Dependent Plasticity is to partition a stream of input patterns into clusters and associate each cluster with an excitatory neuron. The values of a neuron's synaptic weights are highly correlated with the neuron's cluster's centroid.

STDP is implemented as a parallel set of small finite state machines, one at each synapse. As each sequential input is applied, the synaptic weight $w_{ij}$ is updated for the next sequential input using only the synapse's input spike time $x_i(t)$, the output spike time $y_j(t)$ of its associated neuron, and the current value of $w_{ij}$.

Table 1 defines the STDP update function for a given synaptic weight $w$. (Subscripts are omitted to simplify the table). STDP is divided into four major cases, corresponding to the four combinations of input and output spikes being present ($\neq \infty$) or absent ( $= \infty$) . When both are present, two sub-cases are based on the relative timing of the input and output spikes in the classical STDP manner [6].

**Table 1. STDP Update Function**

| input conditions | | weight update |
|---|---|---|
| $x(t) \neq \infty$ | $x(t) \leq y(t)$ | $\Delta w = +\mathrm{B}(\mu_{\text{capture}})*\mathrm{B}(max(\mathrm{F}_{+}(w), \mu_{\min}))$ |
| $y(t) \neq \infty$ | $x(t) > y(t)$ | $\Delta w = -\mathrm{B}(\mu_{\text{backoff}})*\mathrm{B}(max(\mathrm{F}_{-}(w), \mu_{\min}))$ |
| $x(t) \neq \infty$ | $y(t) = \infty$ | $\Delta w = +\mathrm{B}(\mu_{\text{search}})$ |
| $x(t) = \infty$ | $y(t) \neq \infty$ | $\Delta w = -\mathrm{B}(\mu_{\text{backoff}})*\mathrm{B}(max(\mathrm{F}_{-}(w), \mu_{\min}))$ |
| $x(t) = \infty$ | $y(t) = \infty$ | $\Delta w = 0$ |

The "search" mode is new. It occurs when there is an input spike, but no output spike. Glial cells sometimes attach to a synaptic connection, forming a *tripartite* synapse [47]. There is evidence that tripartite synapses support this type of searching behavior. If an input spike repeatedly impinges on the tripartite synapse, the synapse grows in strength even in the absence of neuron output spiking.

The STDP update function either increments the weight by $\Delta w$ (up to a maximum of $w_{\max}$), decrements the weight by $\Delta w$ (down to a minimum of 0), or leaves the weight unchanged. The $\Delta w$ values are defined using Bernoulli random variables (BRVs) with parameterized learning probabilities denoted as $\mu$ with a descriptive subscript. Bernoulli random variables are well-suited to hardware implementations because a linear feedback shift register (LFSR) network can provide pseudo-random binary values.

The STDP update functions include *stabilizing* functions $F_+$ and $F_-$ , described in detail below. For initial discussion, temporarily ignore them (assume $F_+ = F_- = 1$).

Informally, the column-wide STDP strategy follows. First, assume an initial state where all the $w_{ij} = 0$. For the initial input patterns, there will be no output spike. However, weights for all synapses receiving an input spike will be pseudo-randomly incremented by $B(\mu_{search})$. As this stochastic updating continues, the synapses receiving more spikes increment more often. This process of updating synapses in the absence of output spikes is "stochastic search".

Eventually, weights feeding one of the neurons become high enough that an input pattern causes the neuron to fire an output spike. This triggers the "capture" of a cluster: the weights of synapses receiving input spikes will be further increased by $B(\mu_{capture})$, and weights of synapses *not* receiving an input spike will be decremented by $B(\mu_{backoff})$. As a consequence, the neuron's synaptic weights become more strongly associated with the input pattern's cluster.

Meanwhile, input patterns *dissimilar* with respect to the given cluster eventually trigger output spikes in *other* excitatory neurons, and, by the mechanism just described, the other neurons eventually capture clusters.

When in a stable state, every neuron is associated with a cluster. However, the stochastic search mechanism continues in order to adapt to changes in input patterns at the macro-level. As long as a pattern belonging to a cluster appears frequently enough, the backoffs will dominate the searches, and a cluster remains "captured".

However, if the sequence of input patterns *does* change at a macro-level, then a neuron may disassociate itself from one cluster (which has become much less common) and associate itself with a different one (which has become much more common). This occurs if the backoffs don't occur frequently enough to hold back the searches. I.e., stochastic search eventually causes a neuron to fire for, and possibly capture, a new cluster.

The STDP rules used here only depend on the sign of $\Delta t$ and the current weight, as embodied in the stabilization functions $F_+$ and $F_-$ The desired characteristic of the F functions is that if weights are close to 0 or $w_{max}$, they will have a bias to stay there. Following is the definition of function $F_+(w)$; $F_-(w)$ is defined in an analogous manner.

Because $F_+(w)$ is the parameter for a BRV, weight $w$ is first normalized to fall between 0 and 1: $w_n = w/w_{max}$. Then, as a heuristic, $F_+(w) = (w_n)(2\text{-}w_n)$ because it is simple and exhibits the following desired characteristics. If the STDP update rule indicates that $w$ should be incremented and $w$ is close to 0, then $w$ will tend to "stick" close to 0 because the probability of an update is made smaller by a factor of $F_+$. However, if increment conditions occur frequently enough, the rate of increments will increase, and $w$ will gradually accelerate toward $w_{max}$, eventually settling at $w_{max}$.

The F functions are constrained by a floor ($\mu_{min}$) on the values of $F_+$ or $F_-$ . Using $\mu_{min}$ avoids the possibility of a weight becoming permanently stuck at 0 or $w_{max}$. Because the functions are implemented as BRVs, the multiplications implied by Table 1 are implementable as single AND gates.

## 4 Simulations: Decoding Noisy Messages

The following simulations demonstrate operation of the proposed column architecture. In a semi-synthetic demonstration benchmark, a finite set of "messages" or "patterns" are transmitted over a noisy channel. At the receiver, the sequence of noisy received patterns are "de-noised" and decoded to determine the transmitted message.

### 4.1 Simulation Dataset

The simulation dataset is based on ten Arabic numerals selected rather arbitrarily from the 28×28 patterns in the MNIST dataset [33]. These ten baseline patterns were used because they are "organic" in a sense: they are not synthetically contrived to be well-behaved (or ill-behaved) by the researcher.

Long pseudo-random sequences of the ten selected patterns were generated, with pseudo-random noise being added to each. Specifically, a fraction of the pixels are flipped from 1-to-0 or 0-to-1 with probability .30. Preliminary experiments indicated that anything less than .30 is insufficiently challenging.

The ten selected numerals are shown in Figure 5 in binarized form. Most of the simulations reported here use an 8×8 receptive field (RF) taken from the center of the image. A typical noisy version for each of the images is also given.

### 4.2 Evaluation Metrics

The *k-means* algorithm is a standard method for unsupervised clustering. It is a compute-intensive process that begins with a set of *k* randomly or heuristically placed centroids and repeats 2-step epochs: 1) determine clusters based on nearest centroids, 2) determine a new set of centroids based on these clusters. The process continues until the centroids *converge:* centroids at the end of step 2 are the same as (or very similar to) the centroids at the end of step 1.

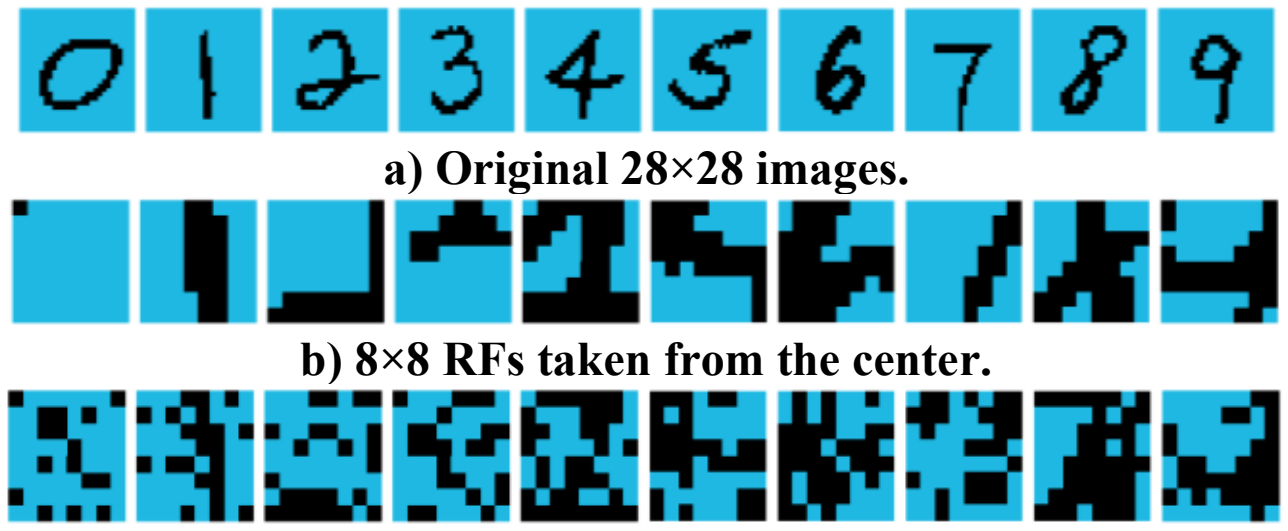

**a) Original 28×28 images.**

**b) 8×8 RFs taken from the center.**

**c)Exampes of noisy 8×8 RFs.**

**Figure 5. Ten baseline numerals selected from the MNIST dataset. The numerals:*image numbers* are: 0:*157* 1:*9* 2:*17* 3:*51* 4:*151* 5:*220* 6:*63* 7:*423* 8:*344* 9:*163*.**

The $k$-means algorithm has an implied internal convergence metric. *Centroid convergence* is the fraction of patterns at the end of an epoch whose nearest centroid is associated with their assigned cluster from the previous epoch. In this paper, the sum of absolute differences (*sad*) distance metric is used, unless stated otherwise. For patterns $x$ and $y$, $sad(x,y) = \Sigma_{i=1..p} \mid x_i - y_i \mid$. Given a set of $q$ clusters $C_{1..q}$ and their associated centroids $\mathbf{c}^{1..q}$, the *nearest centroid* for pattern $x$ is $\mathbf{c}^i$ such that $sad(x, \mathbf{c}^i) \le sad(x, \mathbf{c}^j)$ for all $i \neq j$. If $i$ is not unique, the centroid having the smallest such $i$ is systematically chosen.

$$c_conv = \Sigma_{i=1..q} \Sigma_{x \in C_i} sad(x,\mathbf{c}^i) \le sad(x,\mathbf{c}^j) \,/\, |P| \;,\; i \neq j$$

where $|P| = \Sigma_{i=1..q} |C_i|$.

The *average distance* metric is average distance of input patterns from their cluster's centroid.

$$avg_dist = (\Sigma_{i=1..q} \Sigma_{x \in C_i} \; sad(x,\mathbf{c}^i)) \,/\, |P|$$

*Weight convergence* is a measure of the bimodal character of the synaptic weight distribution. The closer the metric is to 0, the closer all the weights are to either 0 or $w_{\max}$. This metric is adopted from [31].

$$w_conv = [\Sigma_{i=1..q} \Sigma_{j=1..p} \; w_{ij} \, (w_{\max} - w_{ij})] \,/\, p*q*\, w_{\max}$$

For some datasets, input patterns have a known label. After unsupervised training and inference are complete, cluster results can be compared retrospectively with the known labels. *Purity* [3] is a measure of the degree to which all input patterns with the same label are mapped to the same cluster. Assume $l$ labels, and $L_j$ is the set of patterns having label $j$.

$$purity = \Sigma_{i=1..q} \max_{j=1..l} |\, C_i \cap L_j \,| \,/\, |P|$$

If all the elements of each cluster have the same label, then *purity* = 1. For the simulations reported here, purity may be informally interpreted as "accuracy". For a given cluster, the most common label is considered to be "correct", so purity is the fraction of patterns having "correct" labels.

### 4.3 Experimental Configuration

A pre-processing step generates the negative of input images, and both the positive and negative images are applied as column inputs. Hence, an 8×8 image is converted to a *PosNeg* 8×8×2 image that is transmitted to a column as 128 lines, 64 of which contain spikes. In example figures, only the positive part of the image is shown.

This method is similar to (but not the same as) *OnOff* coding in the retinal ganglion cells. Such balanced encoding is important at interfaces where incoming values are not necessarily temporally correlated with strengths.

The excitatory neurons employ ramp no-leak response functions. Training is via STDP as described in Table 1, with $w_{\max} = 8$. Inhibition is 1-WTA with systematic first-index tie breaking. The simulated column is constructed with the foreknowledge that there are 10 clusters, hence there are 10 excitatory neurons. For the 8×8×2 RFs used in the experiments, there are therefore 128 inputs feeding 10 excitatory neurons through a 1280 synaptic crossbar.

The simulator implements an integer model and parameters are specified as non-negative integers. The BRV $\mu$ parameters for STDP updates are specified as an integer numerator, and the denominator is always understood to be 1024. For example if $\mu = 16$, then the BRV $\mu$ parameter =16/1024.

A simulation run consists of 70000 pseudo-random, noisy input patterns. These are simulated with learning always turned on. The first 60000 are considered a *warm-up* sequence, and *test* results are computed only for the final 10000 images.

### 4.4 *k*-means performance

For comparison, the $k$-means algorithm was applied to the benchmark sequence. To be consistent with column simulations, the first 60,000 8×8×2 patterns were used for training. $k$-means clustering was performed for a total of 1024 pseudo-randomly generated sets of initial centroids. For each, the algorithm stopped when the *c_conv* metric reached at least .99. This required 7 to 25 epochs, with 10 being typical. The next 10K noisy input patterns were then partitioned into clusters using the $k$-means centroids, and the purity metric was computed. For the 1024 cases, the best clustering accuracy (as measured by purity) was **.**92; the worst was .74 with a mean of **.**89.

### 4.5 Baseline Results

The first set of simulations began with all weights initialized to 0. A range of search parameters were simulated: $\mu_{\text{search}}$ = 2, 3, 4. For each of these, a parameter space consisting of $\theta$, $\mu_{\text{capture}}$, $\mu_{\text{backoff}}$, and $\mu_{\min}$ was searched, with top performing results chosen.

For a second set of simulations $\mu_{\text{search}} = 0$, and the weights are initialized to relatively high values. The rationale is that with high initial weights, a search mode may not be necessary; the weights are already at a level where any input pattern will produce an output spike, thereby initiating cluster captures. In one simulation run, the weights are initialized to a value of 7, and in the other they are assigned according to a pseudo-random normal distribution with mean = $.80*w_{\max}$ and st. dev. = $.05*\,w_{\max}$. These correspond to the values used in [31].

Overall, there are five STDP configurations plus the $k$-means baseline. Results are in Table 2.

Focus first on the overall performance metrics. They are virtually the same for all five configurations; too similar to justify plotting the results. They all have a purity of .89 to .92. Across the five STDP approaches, the *c_conv* metric is virtually the same, and according to the *w_conv* metric, weights converge to a bimodal distribution as expected. The $k$-means numbers are slightly better than the STDP numbers. Similarity across all the metrics is reassuring -- a similar stable point is reached for a variety of initial conditions and parameters.

**Table 2. Baseline Results**

| | Parameters | | | | Performance Metrics | | | |
|---|---|---|---|---|---|---|---|---|
| **Configuration** | θ | $\mu_{min}$ | $\mu_{capture}$ | $\mu_{backoff}$ | *w_conv* | *avg_dist* | *c_conv* | *purity* |
| $\mu_{search} = 2$ | 60 | 32 | 224 | 304 | 0.01 | 54.5 | 0.94 | 0.90 |
| $\mu_{search} = 3$ | 60 | 32 | 224 | 320 | 0.01 | 54.3 | 0.94 | 0.90 |
| $\mu_{search} = 4$ | 60 | 32 | 224 | 336 | 0.01 | 54.1 | 0.93 | 0.89 |
| $\mu_{search} = 0$, $w = 7$ | 60 | 36 | 256 | 320 | 0.01 | 54.9 | 0.93 | 0.91 |
| $\mu_{search} = 0$, $w = (.8,.05)$ | 60 | 36 | 208 | 304 | 0.01 | 54.9 | 0.93 | 0.89 |
| $k$-means (1024 trials) | | | | | | 53.6 | 0.97 | 0.92 |

### 4.6 STDP Updates

To visualize the convergence process, the counts of STDP update operations for the $\mu_{search} = 3$ configuration are plotted in Figure 6. These are actual weight updates, after the BRVs are applied.

The behavior is as expected. Initially, when the weights start at 0, there is a relatively large number of searches. By the end of the first 1000 inputs, captures and backoffs have kicked-in, indicating the capture of clusters. These peak at 2000 inputs. Searches then become less frequent, and eventually a steady state is achieved where all three update operations reach near-constant levels. The simulated configurations $\mu_{search} = 2,3,4$ have STDP update paths that follow similar phases. In general, the higher the $\mu_{search}$, the faster convergence occurs, but at a cost of more STDP updates.

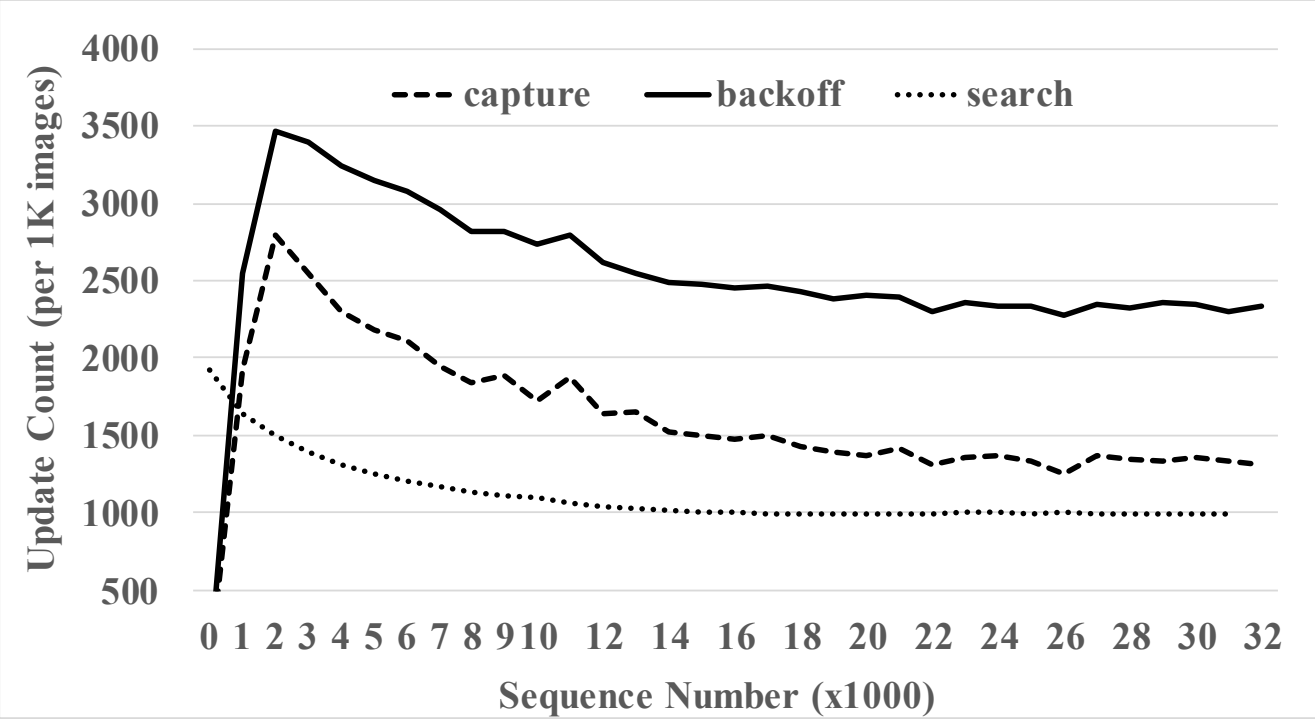

**Figure 6. STDP update counts for $\mu_{search} = 3$.**

The remainder of this subsection is focused on STDP update operations. Convergence is covered in the next subsection. STDP weight update counts are important because they are potentially a big burner of dynamic power. There are typically far more synapses than neuron bodies, so power consumption in the synapses is a major concern.

For five simulated configurations, the average numbers of updates per applied input pattern are given in Figure 7. There are two sets of data: one is an average over the first 10K inputs, while the system is learning, and the other is for the last 10K, when the system is in stable steady state

The totals vary from 3 to ~7 updates per input pattern, with more during the initial learning period than during steady state. That is, only about 3 to 7 out of the 1280 synaptic weights is updated (fewer than 1 percent) per input pattern. This is good. Updates are very sparse, so energy consumption will be relatively low. Better yet, this is the case for dense input spiking patterns; sparse input patterns are likely to consume even less energy for synapse updates. Finally, as expected, the higher search values lead to more STDP update activity.

The $\mu_{search} = 0$ models are more economical with respect to STDP updates. Not only are there no searches, but in steady state, backoffs are significantly reduced because they are not longer needed to offset the searches.

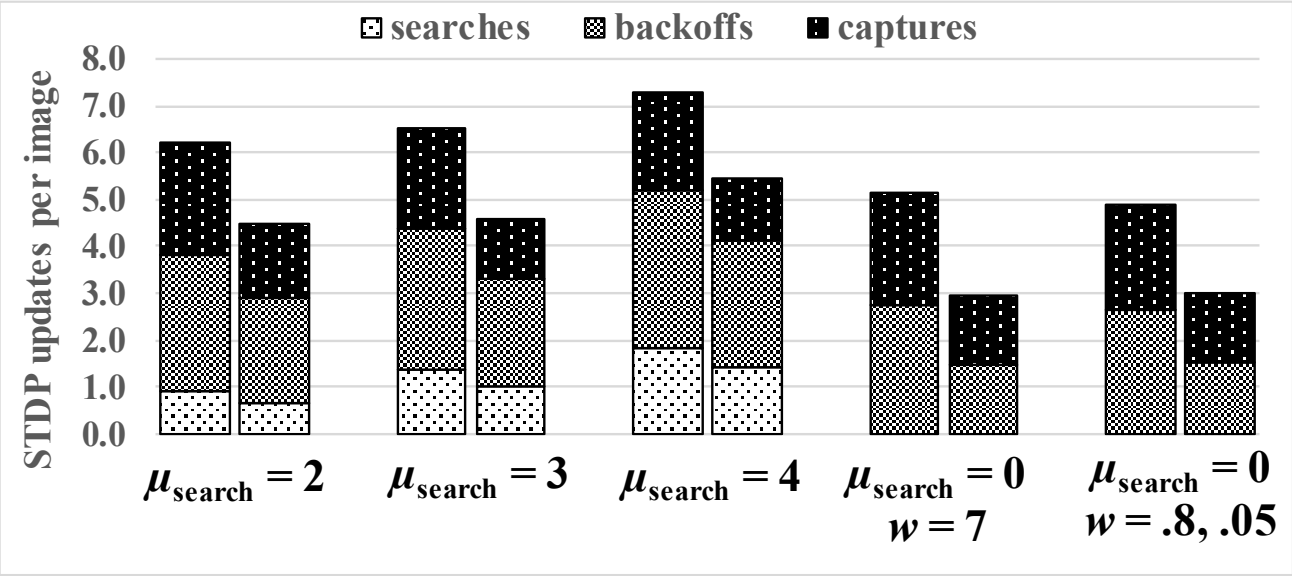

**Figure 7. STDP update counts for five configurations. Left bars are for first 10K inputs; right bars are for last 10K inputs.**

The advantage of *search* is that it enables the STDP mechanism to reach a stable weight distribution regardless of its initial state; the initial state can even be 0, as in the simulations. Another issue is whether the search-less method can easily adapt to macro-level input changes. This is discussed further in Section 4.9.

A good design point for comparison is the STDP update function used in [31]. It implements a version of the $\mu_{search} = 0$ model, where the weights (which range from 0 to 1) are initialized according to a normal distribution with mean = .8 and StD = .05. For STDP updates referred to here as "capture" and "backoff", the functions in [31] are similar to the ones used here.

Performance is on a par with the configurations that implement a search. However, as shown in Section 4.7, convergence to stable weights is slower. And, as noted above, there may be an issue regarding adaptability.

Beyond functional differences, there is also a significant implementation difference between the BRV update approach used here and the much more common deterministic increment approach as exemplified in [31]. In that work, STDP weight increments are deterministically .004 and decrements are -.003. To increment up or decrement down over the full range of weights, from 0 to 1 as in [31], requires at least 300+ separate decrements or 250 increments. Perhaps not a problem in a software implementation, this would be computationally burdensome in a direct hardware implementation as considered here.

For the envisioned direct hardware approach, weights are small integers, and STDP updates are done probabilistically via BRVs. Incrementing up or down the full range of 8 weights requires only 8 updates. Basically, a large number of small updates is being replaced by a much smaller number of large updates (controlled by the BRVs).

### 4.7 Sensitivity and Convergence

Both sensitivity to parameter variations and convergence are studied with the same set of simulations. To illustrate sensitivity to input parameters, the baseline $\mu_{search} = 3$ configuration was simulated over a parameter space consisting of 81 data points; three each for $\theta = 60 \pm 4$; $\mu_{capture} = 224 \pm 16$; $\mu_{backoff} = 336 \pm 16$. The purity results for these 81 parameter sets are sorted from highest to lowest and plotted with solid lines in Figure 8. Best results are for a warm-up period of 60K images, with evaluation over the next 10K ("60K+10K"). This allows a lot of time for stable weights to be reached. As shown, nearly all parameter settings yield a purity that is near the best, so the results are relatively insensitive to a range of learning parameter settings.

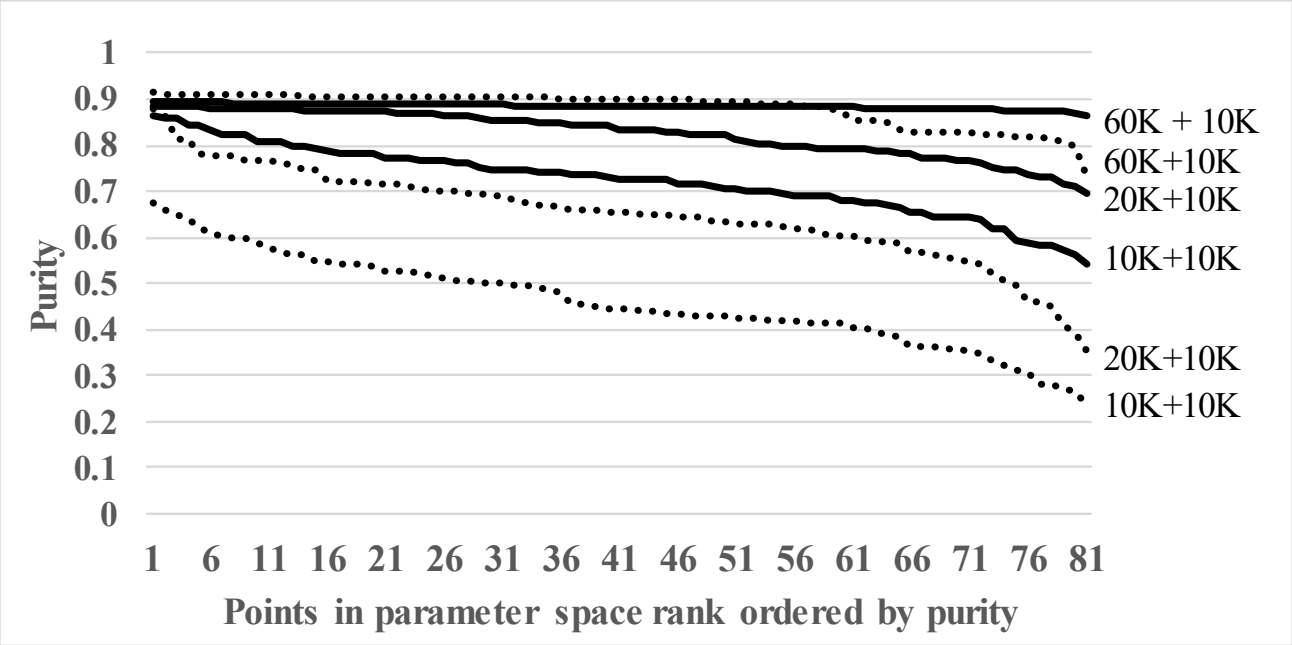

**Figure 8. Sensitivity to parameter variations. Three warm-up periods were used (10K, 20K, 60K) for two configurations: $\mu_{search} = 3$ (solid lines) and $\mu_{search} = 0$, w = 7 (dotted lines).**

Next, consider the other two sets of solid lines in Figure 8: 10K+10K and 20K + 10K. A warm-up of 20K is sufficient for convergence in about a quarter of the cases. With a warm-up of 10K, only a few parameter settings yield passable purity levels.

In contrast, similar simulations with $\mu_{search} = 0$, $w = 7$ are plotted with doted lines. These also perform well with the long 60K warmup period, but not over as wide a parameter space as with $\mu_{search} = 3$. The 20K and 10K warm-ups perform significantly worse than $\mu_{search} = 3$ for all the parameter settings. This is indicative of slower convergence in the absence of a search mechanism.

Combining the observations regarding convergence with the results in Figure 7, the expected tradeoff between learning times and STDP update counts is clearly present. Higher search parameters lead to faster learning, but they also lead to higher STDP update counts. This tradeoff can be managed by adjusting the $\mu_{search}$ parameter (combined with corresponding adjustments in $\mu_{capture}$, $\mu_{backoff}$).

### 4.8 Temporal Processing

An important design objective is achieving "RBF behavior"; i.e., the presence of an output spike not only indicates cluster membership, but the relative spike time indicates the distance from the centroid.

In an idealized system, consider the relationship be number of feature matches, the threshold, and the output spike time. The number of feature matches is indicative of the distance from the centroid: more matches imply smaller distance. In the idealized system, assume the ramp response function does not flatten; I.e., $\rho(w,t) = 1 + t$ ; $t \geq 0$.

With binarized inputs, all ramp responses begin at $t = 0$. For $m$ feature matches, the body potential $v_m$ is the sum of $m$ response functions $\rho$: $v_m(t) = m + m{*}t$ . An output spike is produced at the time the threshold is reached; for $m$ matches, define this to be $t_m$ , so the condition for outputting a spike is: $m + m{*}t_m \geq \theta$ . Because times are integers:

$$t_m = \lceil \theta/m \rceil - 1.$$

This is an inverse relationship: for values close to $m$, relative spike times will be packed closer together. This is illustrated in Figure 4. Body potential for 1 through 4 spikes, where all syn-apses have weight $w_{max} = 8$. In that example, however, the numbers of feature matches are relatively low (4 or fewer), so the spike times offer good temporal resolution. With high numbers of spikes, 64 in the simulated RFs, the threshold must be fairly high in order to get similar temporal resolutions.

For the column under consideration, say satisfactory temporal resolution is about 3 bits, or separating numbers of feature matches into roughly 8 buckets according to output spike times. For the simulated columns, this can be achieved with a threshold of 512 (see Figure 9). In the figure, both cumulative coverage and purity are plotted. In the example plots, the coverage "sweet spot" consists of spike times 14-18. In that region, both cumulative purity and coverage vary approximately linearly, indicating roughly equal size "buckets", with equal size differences in purity. This is very desirable behavior.

*Important design note:* a threshold of 512 is fairly costly if it is implemented in full. However, there is a useful, much cheaper approximation. If a neuron has an implementation threshold $\theta_I$, we can extrapolate the output spike time with a higher functional threshold $\theta_F$.

Under the simplifying assumptions of binarized inputs and unbounded ramp functions, the sum of the response functions is also a linear ramp. Let $z_I$ be the output spike time using the implemented threshold $\theta_I$. If $v$ is the body potential of the neuron when $\theta_I$ is reached, then $z_F/\theta_F = z_I/v$, and because a threshold crossing occurs at an integer time: $z_F = \lceil z_I\, \theta_F / v \rceil$. This approximation is implemented in the simulator that produced results in Figure 9.

Finally, note that the requirement of high thresholds arises because of the input high spike counts. However, with one hot cluster identifiers, a column's output volleys are sparse. And when layered in a TNN, columns other than the first will typically receive sparse input volleys so a low threshold will be sufficient for good RBF behavior.

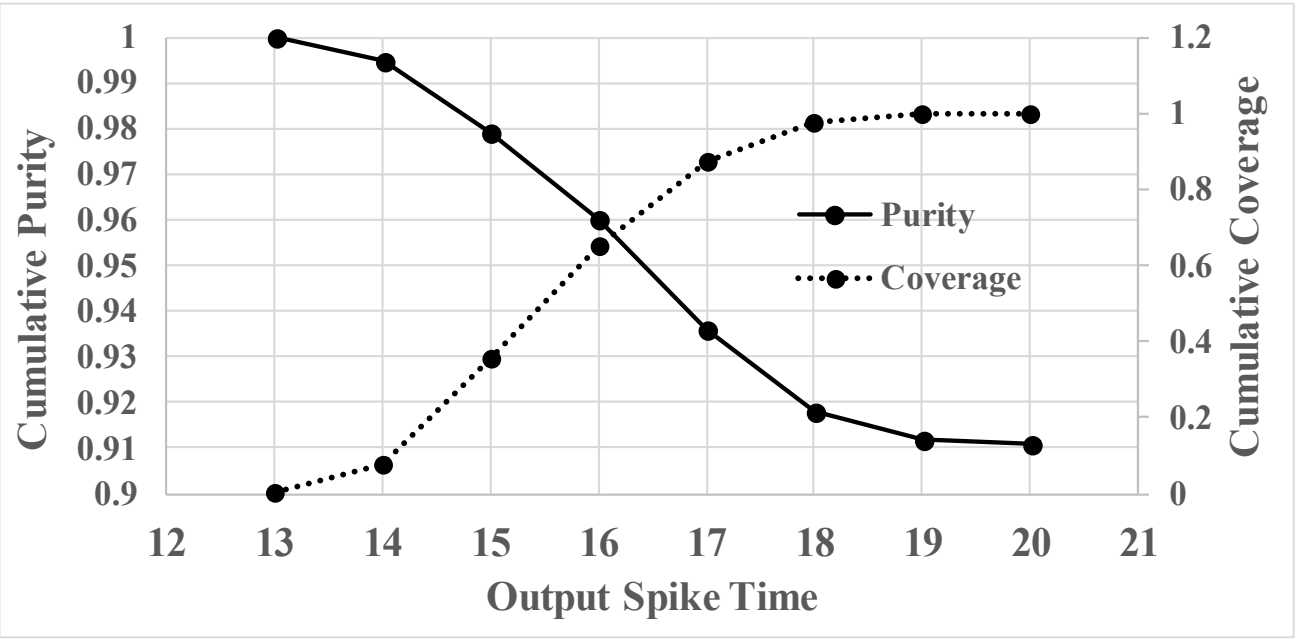

**Figure 9 Cumulative coverage and purity for θ = 512.**

A commonly used response function is step no-leak [31]. If a step no-leak response function is substituted for the ramp no-leak function, then the key metrics do not change: *avg_dist*, *c_conv*, and *purity* are the same. In other words, the step no-leak finds the cluster centroids just as well as the ramp no-leak does. On the other hand, RBF behavior is completely absent. When the input is binarized, all output spikes occur at exactly the same time; there is no temporal processing. So, the ramp no-leak response function provides additional computational capability: RBF behavior.

A classic theoretical result due to Maass [36], is that a step response is functionally less capable than a ramp response function (subject to a set of reasonable assumptions). This result is aligned with the observation just made, and for similar reasons.

### 4.9 Adaptability

A important feature of online learning is that if the input patterns change at the macro level, then the cluster centroids dynamically adapt to the new input patterns. To demonstrate this, an input sequence consisting only of the odd numerals is first applied, followed by a sequence consisting of the even numerals. The transition point happens to fall at 34916. The same 8×8×2 RFs and column architecture are used as before. Both the $\mu_{search}$ = 3 configuration and the $\mu_{search}$ = 0 configuration ($w$ = 7) were simulated. The odd sequence was warmed up for the first 24916 inputs and metrics were computed over the next 10K inputs -- up to the odd-even transition. For the even numerals, warm-up consisted of the first 60K patterns -- through all the odd numerals and past the transition to the even numerals -- and metrics were computed for the final 10K evens.

**Table 3. Odd-Even Results**

| | Parameters | | | | Performance Metrics | | |
|---|---|---|---|---|---|---|---|
| **Configuration** | θ | $\mu_{min}$ | $\mu_{capture}$ | $\mu_{backoff}$ | *w_conv* | *avg_dist* | *purity* |
| $\mu_{search}$ = 3 odds | 60 | 32 | 224 | 320 | 0.01 | 54.1 | 0.90 |
| $\mu_{search}$ = 3 evens | 60 | 32 | 224 | 320 | 0.01 | 57.9 | 0.94 |
| $\mu_{search}$ = 0 $w$ = 7 odds | 60 | 36 | 256 | 320 | 0.01 | 54.4 | 0.91 |
| $\mu_{search}$ = 0 $w$ = 7 evens | 60 | 36 | 256 | 320 | 0.01 | 56.4 | 0.93 |

Results are in Table 3. The important takeaway is that both configurations successfully adapt when odd numerals switch to even numerals. The purity numbers are high, across the board. The *w_conv* metric shows both configurations re-converge to a bimodal weight distribution after the odd-even transition. Figure 10 shows the STDP update profiles for the $\mu_{search}$ = 3 configuration. The transition from odds to evens is very evident: there is an abrupt increase in capture/backoff activity, which eventually settles to stable values. Although not plotted, the $\mu_{search}$ = 0 configuration displays similar behavior.

The successful transition when $\mu_{search}$ = 0 demonstrates that the search function is not required for such a transition. *This was unexpected.* The thought was that in the absence of a search mechanism, high initial weights gradually descend down to stable levels. Once stabilized, the weights may be unable to raise themselves in order to descend to new stable levels.

What actually happens for this data set is the column under study receives inputs with high spike counts. This, coupled with correspondingly dense weights, and a relatively low threshold, means that any novel input will match some existing cluster close enough that an output spike will be produced, and a search is not needed to set the cluster capture process in motion.

However, if both inputs and non-zero weights are sparse, it is not clear that there will always be sufficient spiking activity to initiate the capture/backoff switchover to new clusters. In such cases, a non-zero $\mu_{search}$ parameter may be necessary. This remains to be experimentally determined.

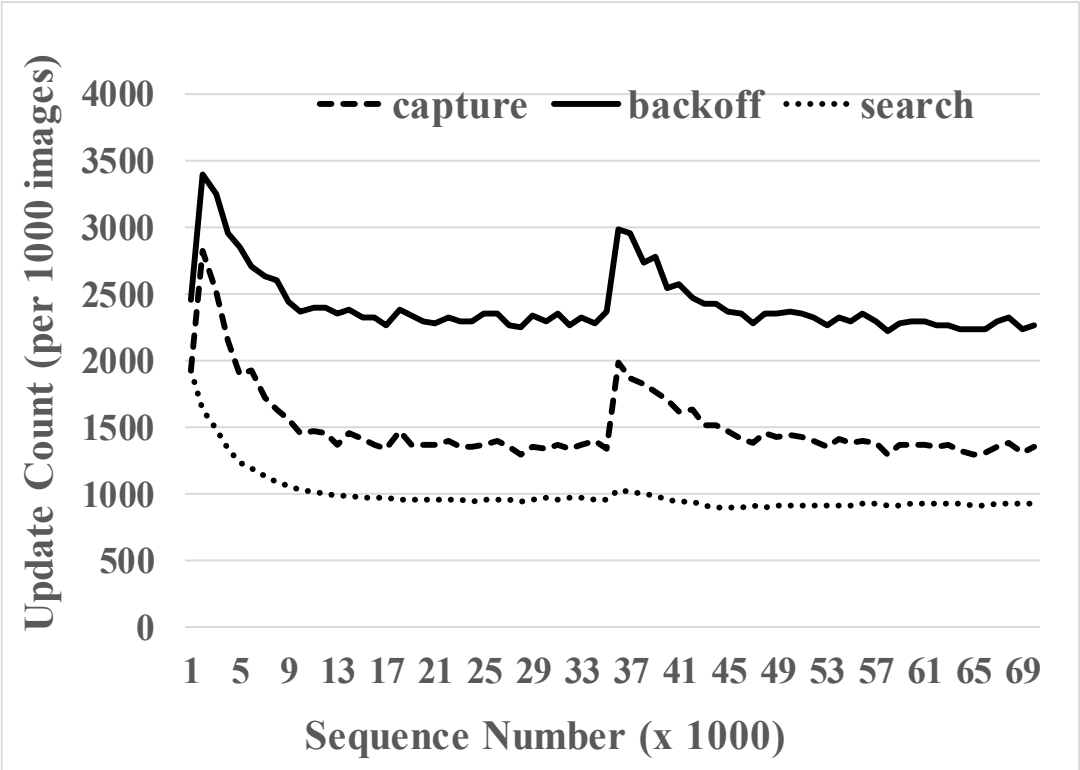

**Figure 10. STDP updates for odd numerals followed by even numerals. $\mu_{search}$ = 3 configuration.**

### 4.10 18×18 Results

To show that a column's clustering capabilities extend beyond 8×8×2 RFs, larger 18×18×2 RFs from the middle of the 28×28 images were applied to a scaled up network, consisting of 10 excitatory neurons fed by 648 synapses. Two column configurations were implemented ($\mu_{search}$ = 0 and $\mu_{search}$ = 4). Performance with $k$-means was also computed. Results are in Table 4. All three methods achieve ideal convergence of 1.0 and purity of 1.0.

**Table 4. Simulation Results for 18 x 18 Images.**

| | Parameters | | | | Performance Metrics | | |
|---|---|---|---|---|---|---|---|
| **Configuration** | θ | $\mu_{min}$ | $\mu_{capture}$ | $\mu_{backoff}$ | *avg_dist* | *c_conv* | *purity* |
| $\mu_{search} = 4$ | 56 | 40 | 232 | 288 | 274 | 1.00 | 1.00 |
| $\mu_{search} = 0$ | 56 | 40 | 224 | 240 | 274 | 1.00 | 1.00 |
| *k*-means | | | | | 272 | 1.00 | 1.00 |

## 5 Summary and Future Research

A neuromorphic column architecture is based on five basic elements: temporal coding, excitatory neuron modeling (inference), WTA inhibition, column structure, and STDP training. This paper describes a prototype column constructed as a combination of these elements. Through experimental simulations of a semi-synthetic benchmark, the column is shown to operate as predicted by the underlying theory. Metrics indicate clustering capability that is on par with classic epoch-based *k*-means clustering. An important difference between the two is that the STDP method proposed here is online, whereas the *k*-means method is offline.

The STDP method, based on BRVs, provides online unsupervised learning with very sparse weight updates, even for dense input spiking patterns. The STDP method studied here adds an always-active "search" mode that is invoked when a synapse receives an input spike, but its associated neuron does not produce an output spike. The search mode reduces learning times and enables synaptic weight convergence regardless of initial weights.

A ramp response function supports temporal processing which is manifested in RBF behavior. This is in contrast to the commonly used step response functions. Using a step response function leads naturally to event-driven implementations -- an output spike can only be triggered at the time an input spike is received. Hence, being "event-driven" is often touted as an SNN advantage. However, the results here demonstrate that some functionality *may* be lost when this is done. In a multi-layer TNN, the strength of cluster matches can be passed from one layer to the next, and this may provide better accuracies. However, to be clear, this remains to be demonstrated via experimentation, as only a single column was studied in this paper. A peripheral argument in favor of a ramp response function appears in theoretical work by Maass [36] where ramp response functions are shown to be more powerful than step response functions.

This research is part of an ongoing effort targeting a direct hardware implementation: a silicon neocortex. This is admittedly ambitious, but is driven by the working hypothesis that much of the brain's efficiency comes from using the flow of time as a resource in the implementation. To facilitate direct hardware implementation, the model column operates on low precision values encoded as spikes, or, more generally, transients in time.

Regarding future research, at the forefront is the development of column architectures for a variety of edge processing applications: denoising, lossy compression, front-end ML processing, anomaly detection.

At the lower level, silicon implementations and related analysis of area, energy, and delay can shed significant light on the performance and efficiency of a future silicon neocortex.

With regard to TNN architectures, a next step is to organize multiple columns in a hierarchical, layered structure to form a general TNN that parallel processes inputs from many RFs. Then, temporally encoded information from the first layer, implemented as described here, becomes temporal input into the second layer. Of special research interest are the ways a second layer can exploit its RBF-encoded inputs. Furthermore, the second layer will be a good testbed for input volleys that are much sparser than the dense inputs considered in this paper.

## 6 References

[1] Abeles M. *Local Cortical Circuits: An Electrophysiological study.* Springer, Berlin, (1982).

[2] Abeles, Moshe. "Synfire chains." *Scholarpedia* 4, no. 7 (2009): 1441.

[3] Amigó, Enrique, Julio Gonzalo, Javier Artiles, and Felisa Verdejo. "A comparison of extrinsic clustering evaluation metrics based on formal constraints." *Information retrieval* 12, no. 4 (2009): 461-486.

[4] R. Ananthanarayanan1, S. K. Esser, H. D. Simon, and D. S. Modha, "Cognitive computing building block: A versatile and efficient digital neuron model for neurosynaptic cores." *High Performance Computing Networking, Storage and Analysis*. IEEE, (2009).

[5] Beyeler, Michael, Nikil D. Dutt, and Jeffrey L. Krichmar. "Categorization and decision-making in a neurobiologically plausible spiking network using a STDP-like learning rule." *Neural Networks* 48 (2013): 109-124.

[6] Bi, Guo-qiang, and Mu-ming Poo. "Synaptic modifications in cultured hippocampal neurons: dependence on spike timing, synaptic strength, and postsynaptic cell type." *The Journal of neuroscience* 18, no. 24 (1998): 10464-10472.

[7] Bichler, Olivier, Damien Querlioz, Simon J. Thorpe, Jean-Philippe Bourgoin, and Christian Gamrat. "Extraction of temporally correlated features from dynamic vision sensors with spike-timing-dependent plasticity." *Neural Networks* 32 (2012): 339-348.

[8] Bohte, Sander M., Han La Poutré, and Joost N. Kok. "Unsupervised clustering with spiking neurons by sparse temporal coding and multilayer RBF networks," *IEEE Transactions on Neural Networks,* 13, no. 2 (2002): 426-435.

[9] Bohte, Sander M., Joost N. Kok, and Han La Poutre. "Error-backpropagation in temporally encoded networks of spiking neurons." *Neurocomputing* 48, no. 1 (2002): 17-37.

[10] Brader, Joseph M., Walter Senn, and Stefano Fusi. "Learning real-world stimuli in a neural network with spike-driven synaptic dynamics." *Neural computation* 19, no. 11 (2007): 2881-2912.

[11] Butts, Daniel A., Chong Weng, Jianzhong Jin, Chun-I. Yeh, Nicholas A. Lesica, Jose-Manuel Alonso, and Garrett B. Stanley. "Temporal precision in the neural code and the timescales of natural vision." *Nature* 449, no. 7158 (2007): 92-95.

[12] Cao, Yongqiang, Yang Chen, and Deepak Khosla. "Spiking deep convolutional neural networks for energy-efficient object recognition." *International Journal of Computer Vision* 113, no. 1 (2015): 54-66.

[13] Cassidy, A. S., P. Merolla, J. V. Arthur, S. K. Esser, B. Jackson, R. Alvarez-Icaza, P. Datta, J. Sawaday, T. M. Wong, V. Feldman, A. Amir, D. B.-D. Rubinx, F. Akopyan, E. McQuinn, W. P. Risk, and D. S. Modha "Cognitive computing building block: A versatile and efficient digital neuron model for neurosynaptic cores." *International Joint Conference on Neural Networks (IJCNN). IEEE*. 2013.

[14] Comsa, Iulia M., Krzysztof Potempa, Luca Versari, Thomas Fischbacher, Andrea Gesmundo, and Jyrki Alakuijala. "Temporal coding in spiking neural networks with alpha synaptic function." *arXiv preprint arXiv:1907.13223* (2019).

[15] Davies, Mike, Narayan Srinivasa, Tsung-Han Lin, Gautham Chinya, Yongqiang Cao, Sri Harsha Choday, Georgios Dimou et al. "Loihi: A neuromorphic manycore processor with on-chip learning." *IEEE Micro* 38, no. 1 (2018): 82-99.

[16] Diamond, Alan, Michael Schmuker, and Thomas Nowotny. "An unsupervised neuromorphic clustering algorithm." *Biological cybernetics* (2019): 1-15.

[17] Diehl, Peter U., and Matthew Cook. "Unsupervised learning of digit recognition using spike-timing-dependent plasticity." *Frontiers in computational neuroscience* 9 (2015).

[18] Dong M, Huang X, Xu B (2018), "Unsupervised speech recognition through spiketiming dependent plasticity in a convolutional spiking neural network," PLoS ONE 13(11): e0204596.

[19] Ferré, Paul et al. "Unsupervised Feature Learning With Winner-Takes-All Based STDP" *Frontiers in computational neuroscience* vol. 12 24. 5 Apr. 2018, doi: 10.3389 /fncom.2018.00024.

[20] Fusi, Stefano, and Maurizio Mattia. "Collective behavior of networks with linear (VLSI) integrate-and-fire neurons." *Neural Computation* 11.3 (1998): 633-652.

[21] Gerstner, Wulfram, and J. Leo Van Hemmen. "How to describe neuronal activity: spikes, rates, or assemblies?" In *Advances in neural information processing systems*, (1993): 463-470.

[22] Gerstner, Wulfram, Richard Kempter, J. Leo van Hemmen, and Hermann Wagner. "A neuronal learning rule for sub-millisecond temporal coding." *Nature* 383, no. 6595 (1996): 76-78.

[23] Gray, Charles M., Peter König, Andreas K. Engel, and Wolf Singer. "Oscillatory responses in cat visual cortex exhibit inter-columnar synchronization which reflects global stimulus properties." *Nature* 338, no. 6213 (1989): 334.

[24] Gütig, Robert, and Haim Sompolinsky. "The tempotron: a neuron that learns spike timing–based decisions." Nature neuroscience 9, no. 3 (2006): 420-428.

[25] Guyonneau, Rudy, Rufin Vanrullen, and Simon J. Thorpe. "Neurons tune to the earliest spikes through STDP." *Neural Computation* 17, no. 4 (2005): 859-879.

[26] Hebb, D.O. *The Organization of Behavior.* New York: Wiley & Sons, 1949.

[27] Hodgkin, Alan L., and Andrew F. Huxley, A Quantitative Description of Membrane Current and Its Application to Conduction and Excitation in Nerve -- *The Journal of physiology* 117.4 (1952): 500.

[28] Hopfield, J. J. "Pattern recognition computation using action potential timing for stimulus representation." *Nature* 376 (1995): 33.

[29] Hunsberger, Eric, and Chris Eliasmith. "Spiking deep networks with LIF neurons." *arXiv preprint arXiv:1510.08829* (2015).

[30] Karnani, Mahesh M., Masakazu Agetsuma, and Rafael Yuste. "A blanket of inhibition: functional inferences from dense inhibitory connectivity." *Current opinion in neurobiology* 26 (2014): 96-102.m

[31] Kheradpisheh, Saeed Reza, Mohammad Ganjtabesh, Simon J. Thorpe, and Timothée Masquelier. "STDP-based spiking deep neural networks for object recognition." *Neural Networks* 99 (2018): 56-67.

[32] Kim, Seijoon, Seongsik Park, Byunggook Na, and Sungroh Yoon. "Spiking-yolo: Spiking neural network for real-time object detection." *arXiv preprint arXiv:1903.06530* (2019).

[33] LeCun, Yann, Corinna Cortes, and C. J. Burges. "MNIST handwritten digit database." (2010): 18.

[34] Lee, Jun Haeng, Tobi Delbruck, and Michael Pfeiffer. "Training Deep Spiking Neural Networks using Backpropagation." *Frontiers in neuroscience* 10 (2016): 508.

[35] Levy, W. B., and O. Steward. "Temporal contiguity requirements for long-term associative potentiation/ depression in the hippocampus." *Neuroscience* 8, no. 4 (1983): 791-797.

[36] Maass, Wolfgang, "Networks of spiking neurons: the third generation of neural network models." *Neural networks* 10.9 (1997): 1659-1671.

[37] Mainen, Zachary F., and Terrence J. Sejnowski. "Reliability of spike timing in neocortical neurons." *Science* 268, no. 5216 (1995): 1503-1506.

[38] Majani, E., Ruth Erlanson, and Yaser S. Abu-Mostafa. "On the k-winners-take-all network." *Advances in neural information processing systems* (1989): 634-642.

[39] Markram, Henry, Joachim Lübke, Michael Frotscher, and Bert Sakmann. "Regulation of synaptic efficacy by coincidence of postsynaptic APs and EPSPs." *Science* 275, no. 5297 (1997): 213-215.

[40] Masquelier, Timothée, and Simon J. Thorpe. "Unsupervised learning of visual features through spike timing dependent plasticity." *PLoS Comput Biol* 3, no. 2 (2007): e31.

[41] Meftah, B., O. Lezoray, and A. Benyettou. "Segmentation and Edge Detection Based on Spiking Neural Network Model." *Neural Processing Letters* 32, no. 2 (2010): 131-146.

[42] Mostafa, Hesham. “Supervised learning based on temporal coding in spiking neural networks.” *IEEE trans. on neural networks and learning systems* 29, no. 7 (2018): 3227-3235.

[43] Mountcastle, Vernon B. “The columnar organization of the neocortex.” *Brain* 120, no. 4 (1997): 701-722.

[44] Natschläger, Thomas, and Berthold Ruf. “Spatial and temporal pattern analysis via spiking neurons.” *Network: Computation in Neural Systems* 9, no. 3 (1998): 319-332.

[45] Neftci, Emre O., Charles Augustine, Somnath Paul, and Georgios Detorakis. “Event-driven random back-propagation: Enabling neuromorphic deep learning machines.” *Frontiers in neuroscience* 11 (2017): 324.

[46] Nessler, Bernhard, Michael Pfeiffer, and Wolfgang Maass. “STDP enables spiking neurons to detect hidden causes of their inputs.” *NIPS* (2009): 1357-1365.

[47] Perea, Gertrudis, Marta Navarrete, and Alfonso Araque. "Tripartite synapses: astrocytes process and control synaptic information." *Trends in neurosciences* 32, no. 8 (2009): 421-431.

[48] Panda, Priyadarshini, and Kaushik Roy. “Unsupervised regenerative learning of hierarchical features in spiking deep networks for object recognition.” *2016 International Joint Conference on Neural Networks* (2016): 299-306.

[49] Pérez Carrasco, José Antonio, Bo Zhao, María del Carmen Serrano Gotarredona, Begoña Acha, Teresa Serrano Gotarredona, Shouchun Cheng, and Bernabé Linares Barranco. “Mapping from Frame-Driven to Frame-Free Event-Driven Vision Systems by Low-Rate Rate-Coding and Coincidence Processing. Application to Feed-Forward ConvNets.” *IEEE transactions on pattern analysis and machine intelligence,* 35 11 (2013): 2706-2719.

[50] Sengupta, Abhronil, Yuting Ye, Robert Wang, Chiao Liu, and Kaushik Roy. “Going Deeper in Spiking Neural Networks: VGG and Residual Architectures.” *Frontiers in neuroscience* 13 (2019).

[51] Sporea, Ioana, and André Grüning. “Supervised learning in multilayer spiking neural networks.” *Neural computation* 25, no. 2 (2013): 473-509.

[52] Stein, Richard B. “A Theoretical Analysis of Neuronal Variability.” *Biophysical Journal* 5.2 (1965): 173-194.

[53] Tavanaei, Amirhossein & Ghodrati, Masoud & Kheradpisheh, Saeed Reza & Masquelier, Timothée & Maida, Anthony. “Deep Learning in Spiking Neural Networks.” *Neural Networks* 111 (2019): 47-63.

[54] Tavanaei, Amirhossein, and Anthony Maida. “BP-STDP: Approximating backpropagation using spike timing dependent plasticity.” *Neurocomputing* 330 (2019): 39-47.

[55] Thorpe, Simon J., and Michel Imbert. “Biological constraints on connectionist modelling.” *Connectionism in perspective* (1989): 63-92.

[56] Thorpe, Simon J. “Spike arrival times: A highly efficient coding scheme for neural networks.” *Parallel processing in neural systems* (1990): 91-94.

[57] Tuckwell, Henry C. “Synaptic transmission in a model for stochastic neural activity.” *Journal of theoretical biology* 77.1 (1979): 65-81.

[58] Van Rullen, Rufin, Jacques Gautrais, Arnaud Delorme, and Simon Thorpe. “Face processing using one spike per neurone.” *BioSystems* 48 (1998): 229-239.

[59] U. Weidenbacher and H. Neumann, “Unsupervised learning of head pose through spike-timing dependent plasticity,” *Perception in Multimodal Dialogue Systems, ser. Lecture Notes in Computer Science*. Springer Berlin 1 Heidelberg, vol. 5078/2008 (2008): 123-131.

[60] Wysoski, Simei Gomes, Lubica Benuskova, and Nikola Kasabov. “Evolving spiking neural networks for audiovisual information processing.” *Neural Networks* 23 (2010): 819-835.

[61] Zhao, Bo, Ruoxi Ding, Shoushun Chen, Bernabe Linares-Barranco, and Huajin Tang. “Feedforward categorization on AER motion events using cortex-like features in a spiking neural network.” *IEEE transactions on neural networks and learning systems* 26, no. 9 (2015): 1963-1978.